\documentclass[journal]{IEEEtran}
\usepackage{flushend}
\usepackage{amsmath,amsfonts,amsmath}
\usepackage{algorithmic}
\usepackage{algorithm}
\usepackage{array}
\usepackage[caption=false,font=normalsize,labelfont=sf,textfont=sf]{subfig}
\usepackage{textcomp}
\usepackage{graphicx}
\usepackage{amssymb}
\usepackage{booktabs}
\usepackage{multirow}
\usepackage{color}
\usepackage{stfloats}
\usepackage{url}
\usepackage{verbatim}
\usepackage{graphicx}
\usepackage{cite}
\usepackage{bbm}
\usepackage[pagebackref,breaklinks,colorlinks]{hyperref}
\usepackage{xcolor}
\usepackage[capitalize]{cleveref}
\usepackage{colortbl}
\usepackage{marvosym}
\Crefname{section}{Section}{Sections}
\Crefname{table}{Table}{Tables}
\Crefname{figure}{Figure}{Figures}
\usepackage{marvosym}

\begin{document}

\title{Transforming Image Super-Resolution: \\A ConvFormer-based Efficient Approach}

\author{Gang~Wu,        
        Junjun~Jiang\textsuperscript{\Letter},~\IEEEmembership{Senior Member,~IEEE},
        Junpeng~Jiang,
        and Xianming Liu,~\IEEEmembership{Member,~IEEE}

\IEEEcompsocitemizethanks{

\IEEEcompsocthanksitem G. Wu, J. Jiang, J. Jiang, and X. Liu are with the School of Computer Science and Technology, Harbin Institute of Technology, Harbin 150001, China. E-mail: \{gwu@hit.edu.cn, 
jiangjunjun@hit.edu.cn, 1190200226@stu.hit.edu.cn, csxm@hit.edu.cn\}. Corresponding author: Junjun Jiang.

}
}
\markboth{Accepted to IEEE Transactions on Image Processing}
 {Shell \MakeLowercase{\textit{et al.}}: A Sample Article Using IEEEtran.cls for IEEE Journals}

\maketitle

\begin{abstract}
Recent progress in single-image super-resolution (SISR) has achieved remarkable performance, yet the computational costs of these methods remain a challenge for deployment on resource-constrained devices. In particular, transformer-based methods, which leverage self-attention mechanisms, have led to significant breakthroughs but also introduce substantial computational costs. To tackle this issue, we introduce the Convolutional Transformer layer (ConvFormer) and propose a ConvFormer-based Super-Resolution network (CFSR), offering an effective and efficient solution for lightweight image super-resolution. The proposed method inherits the advantages of both convolution-based and transformer-based approaches. Specifically, CFSR utilizes large kernel convolutions as a feature mixer to replace the self-attention module, efficiently modeling long-range dependencies and extensive receptive fields with minimal computational overhead. Furthermore, we propose an edge-preserving feed-forward network (EFN) designed to achieve local feature aggregation while effectively preserving high-frequency information. Extensive experiments demonstrate that CFSR strikes an optimal balance between computational cost and performance compared to existing lightweight SR methods. When benchmarked against state-of-the-art methods such as ShuffleMixer, the proposed CFSR achieves a gain of 0.39 dB on the Urban100 dataset for the x2 super-resolution task while requiring 26\% and 31\% fewer parameters and FLOPs, respectively. The code and pre-trained models are available at \url{https://github.com/Aitical/CFSR}.
\end{abstract}


\section{Introduction}

\IEEEPARstart{S}{ingle} Image Super-Resolution (SISR) is a fundamental task in computer vision that aims to enhance the resolution and quality of a low-resolution image, generating a higher-resolution image with finer details and improved visual quality \cite{survey_TMM,21survey}. The need for SISR arises in various real-world scenarios where high-resolution images are desired but are limited by hardware capabilities or constraints. In many applications, such as surveillance systems, medical imaging, satellite imagery, and digital photography—acquiring high-resolution images may be costly, time-consuming, or restricted. Therefore, SISR techniques provide a valuable solution by leveraging advanced algorithms to upsample low-resolution images.
\begin{figure}[tbp]
\centering
\includegraphics[width=0.475\textwidth]{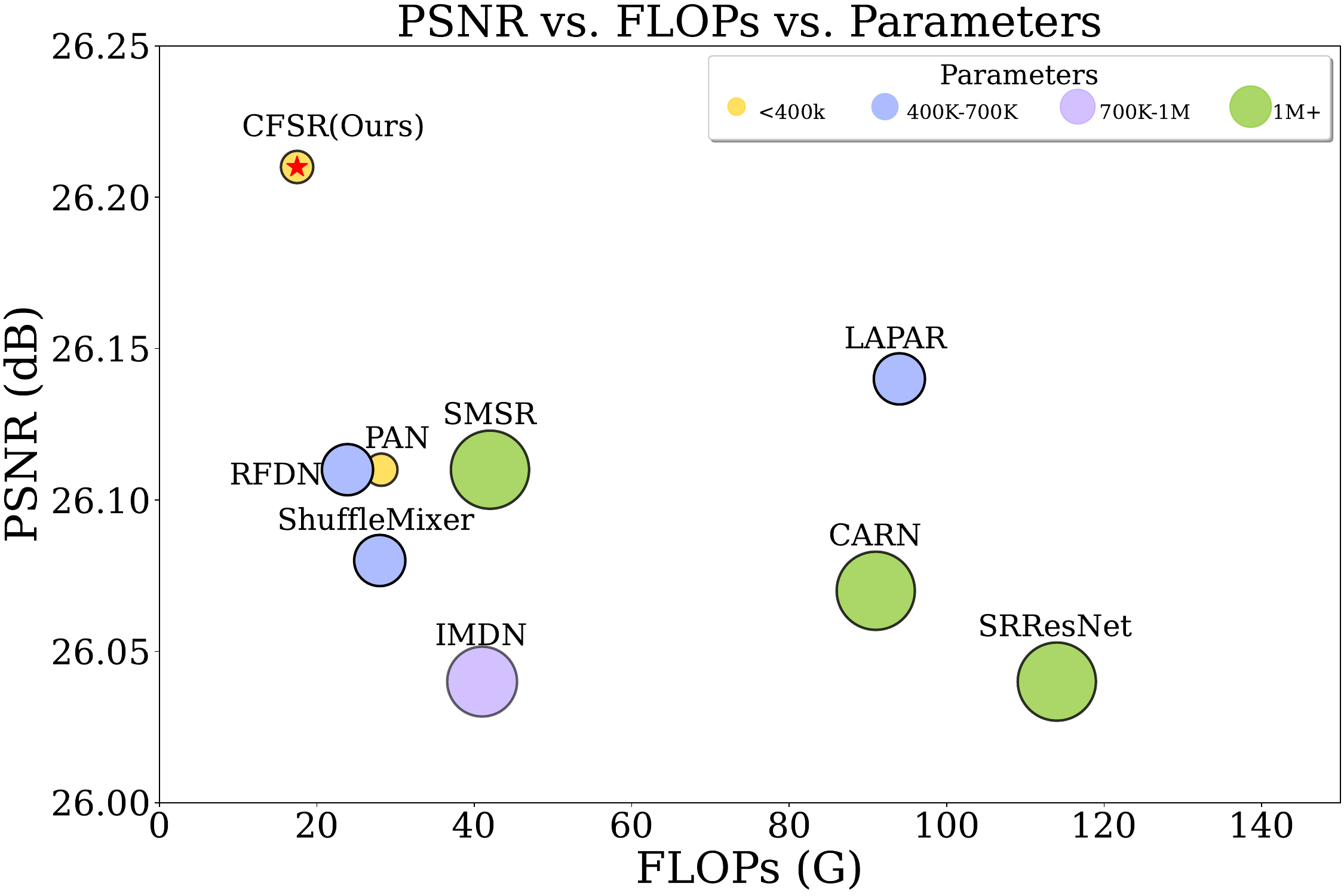}
\vspace{-0.3cm}
\caption{Illustration of PSNR, FLOPs, and parameter counts of different SISR models on the Urban100 dataset for $4\times$ SR task. The proposed CFSR approach achieves superior performance with less computational cost.}
\label{fig:intro_compare}
\end{figure}

In recent years, there has been considerable progress in SISR, largely attributed to the advent of deep learning techniques \cite{CNN-survey,21survey}. A groundbreaking study, SRCNN \cite{SRCNN}, introduced the concept of learning the mapping between low- and high-resolution images using convolutional neural networks (CNNs). This surpassed the performance of previous methods and led to further exploration of CNN-based approaches. Subsequently, numerous studies have developed innovative SR models with deep and effective backbones \cite{VDSR,EDSR} and attention mechanisms \cite{RCAN,SAN,HAN,NLSN}. These CNN-based methods have significantly advanced the state-of-the-art in SISR, demonstrating the power of deep learning in learning complex image representations and generating visually appealing high-resolution images from low-resolution inputs. However, despite the performance gains, these methods typically involve more complex models and higher computational complexity, which hinders their deployment on mobile and edge devices. To address this issue, designing efficient and lightweight super-resolution models has become crucial. Many works have been proposed to reduce the number of parameters or floating point operations (FLOPs) to achieve lightweight models \cite{FSRCNN,CARN,IMDN,LAPAR,SMSR,PAN,ECBSR,FDIWN,shufflemixer}. Zhao \textit{et al.} \cite{PAN} proposed the lightweight pixel-attention network (PAN), which replaces standard residual or dense blocks with an efficient pixel-attention block. Sun \textit{et al.} \cite{shufflemixer} proposed ShuffleMixer, which introduces large kernel convolutions into the lightweight SR network and significantly reduces model complexity through a channel split-shuffle operation. Transformer-based architectures have attracted great attention due to their impressive performance \cite{ViT,SwinT}. The self-attention mechanism provides promising long-range modeling capabilities and has achieved significant breakthroughs in computer vision \cite{ViT}. However, its complexity is quadratic in image size, requiring heavy computational resources. Liu \textit{et al.} \cite{SwinT} proposed the Swin Transformer, which performs self-attention within large local windows. Subsequently, many transformer-based SR methods have been developed \cite{EDT,SwinIR,ELAN,HAT,CAT,GRL}. By leveraging self-attention mechanisms and hierarchical architectures, these methods can effectively capture long-range dependencies and fine details. Transformer-based SR models have often surpassed CNN-based models, achieving new state-of-the-art results in addressing image super-resolution challenges. However, their application in lightweight models is limited due to the high computational cost and substantial CPU or GPU memory requirements of the self-attention mechanism.

To address the challenge of computational efficiency in image super-resolution, we introduce a novel, self-attention-free approach that offers an excellent balance between computational cost and performance, making it a viable solution for practical applications in lightweight image super-resolution. Specifically, we propose the Convolutional Transformer layer (ConvFormer) as a core component for effective and efficient feature extraction. Building on this foundation, we introduce the ConvFormer-based Super-Resolution network (CFSR) tailored for lightweight SISR tasks. As in standard transformer architectures \cite{ViT, SwinT, MetaFormer}, our approach includes a feature mixer module and a feed-forward network. Drawing inspiration from recent successes of CNN-based methods \cite{Conv2Former, VAN, largekernel, convnet}, our proposed feature mixer module employs large kernel convolutions as gating layers. This innovative design eliminates the need for self-attention in the feature mixer module, efficiently capturing long-range dependencies and extensive receptive fields with minimal additional computational cost. Additionally, we introduce an edge-preserving feed-forward network (EFN) that refines the standard feed-forward network by incorporating enhanced edge extraction capabilities. Unlike the conventional feed-forward network (FFN) \cite{ViT}, which incorporates $3\times3$ depth-wise convolutions for improved local feature aggregation in vision tasks \cite{Uformer, Restormer}, our EFN integrates image gradient priors. This integration not only preserves high-frequency information but also introduces significant improvements for lightweight models without increasing complexity or parameter counts during inference, achieved through re-parameterization \cite{RepVGG, ACNet, ECBSR}. The architecture of CFSR, though straightforward and predominantly convolutional, is significantly more effective than previous methods. When benchmarked against existing methods for the $\times4$ SR task on the Urban100 dataset, as detailed in Fig. \ref{fig:intro_compare}, CFSR demonstrates superior performance. It excels in balancing reconstruction quality, model size, and computational efficiency, outperforming state-of-the-art methods with fewer parameters and reduced FLOPs.

We summarize the main contributions of our work as follows:

\begin{enumerate} 
\item We introduce ConvFormer, a feature mixer based on large kernel convolutions that replaces the self-attention module in traditional SISR models. ConvFormer efficiently captures long-range dependencies and extensive receptive fields while maintaining lower computational complexity. This approach demonstrates superior performance and efficiency in lightweight image super-resolution tasks, shedding new light on the design of CNN-based or hierarchical architectures for lightweight SISR. \item We propose EFN, an edge-preserving feed-forward network, to address the loss of high-frequency information in conventional SISR algorithms. EFN incorporates local feature aggregation through convolutional layers with edge-preserving filters and preserves high-frequency information using skip connections and deconvolution layers. Compared to traditional methods, EFN achieves better preservation of image details and textures while maintaining high super-resolution performance. 
\item We present extensive experiments to verify the effectiveness of CFSR. Compared to existing advanced methods, CFSR achieves superior performance with less computational cost. Detailed ablation studies are provided to analyze the impact of different components. Notably, our method outperforms recent NTIRE efficient super-resolution challenge winners. \end{enumerate}

In the following section, we will first give some related work of lightweight image super-resolution methods and the progress of modern architectures in Sec. \ref{sec:related_work}. In Sec. \ref{sec:method}, we introduce and explain our proposed CFSR method in detail. Then, Sec. \ref{sec:experiments} describes our training settings and experimental results including ablation analysis, where we compare the performance of our approach to other state-of-the-art methods. Finally, some conclusions are drawn in Sec. \ref{sec:conclusion}.

\begin{figure*}[htbp]
\centering
\includegraphics[width=0.84\textwidth]{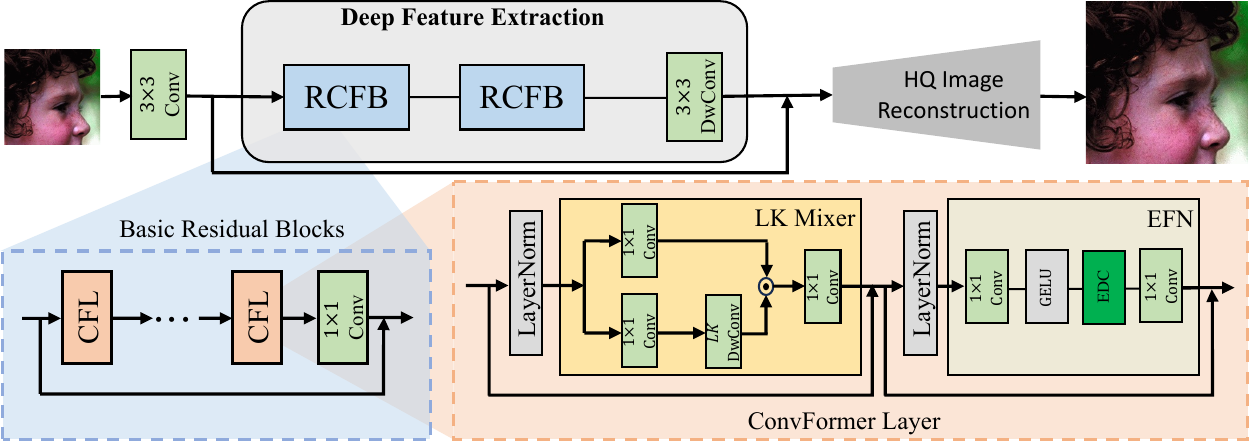}
\vspace{-0.13cm}
\caption{Detailed implementation and different components in the proposed CFSR. The architecture of CFSR is mainly stacked by the basic residual block, which contains several ConvFormer layers. The ConvFormer Block plays a pivotal role, containing the proposed large kernel feature mixer (LK Mixer) and edge-preserving feed-forward network (EFN).}
\label{fig:first}
\end{figure*}

\section{Related Work \label{sec:related_work}}
In this section, we briefly review the related literature, including deep learning-based single-image super-resolution methods, transformer-based architectures, modern convolutional architectures, and the development of re-parameterization methods.

\subsection{Single Image Super-Resolution}

Recently, deep learning methods have achieved dramatic improvements in SISR tasks \cite{survey,ACMComputingSurvey,21survey}. In particular, various well-designed CNN architectures have been explored to further improve SISR performance \cite{VDSR,EDSR,RDN,DBPN,RDN,DRRN}. VDSR \cite{VDSR} introduced a very deep backbone to predict the residual between the low-resolution (LR) input and the corresponding high-resolution (HR) image. EDSR \cite{EDSR} incorporated residual blocks with skip connections, allowing direct propagation of information from earlier layers to later layers. Furthermore, attention mechanisms like channel attention \cite{channel_attention} have been introduced to the SISR task \cite{RCAN,SAN,HAN,NLSN}. Zhang \textit{et al.} \cite{RCAN} proposed the RCAN model, which extends the backbone to over 400 layers by incorporating channel attention mechanisms.

In contrast to achieving advanced performance with a rapidly increased number of parameters and computational cost, many lightweight SISR models have been developed to reduce parameters, especially for resource-limited devices \cite{FSRCNN,CARN,IDN,IMDN,LAPAR,SMSR,ECBSR,FDIWN}. Hui \textit{et al.} proposed the Deep Information Distillation Network (IDN) \cite{IDN} and extended it to the Information Multi-Distillation Network (IMDN) \cite{IMDN}, which won the AIM2020 challenge. Zhang \textit{et al.} \cite{ECBSR} proposed the Edge-oriented Convolution Block (ECB) for real-time inference, which extracts first-order and second-order spatial derivatives from intermediate features.

Recent NTIRE challenges have driven significant advancements in efficient super-resolution. Liu \textit{et al.} introduced the Residual Local Feature Network (RLFN) \cite{Kong2022ResidualLF} in NTIRE 2022, employing large kernel convolutions and deformable convolutions to efficiently capture long-range dependencies. This approach demonstrated the potential of adapting techniques from medical image segmentation to the SR task. In NTIRE 2023, Xu \textit{et al.} proposed the Deep Image Prior Network (DIPNet) \cite{yu2023dipnet}, introducing a multi-stage lightweight network boosting method. DIPNet leverages enhanced high-resolution output as additional supervision and employs network simplification techniques like re-parameterization and iterative pruning.

Most recently, Sun \textit{et al.} \cite{shufflemixer} introduced large kernel convolutions into lightweight SR and proposed ShuffleMixer. Through a channel split-shuffle operation, it efficiently reduces latent projection features. Building on this trend, Zhang \textit{et al.} developed the Swift Parameter-free Attention Network (SPAN) \cite{wan2024swift} for NTIRE 2024, which introduces a novel parameter-free attention mechanism. SPAN uses symmetric activation functions and residual connections to enhance high-contribution information while suppressing redundant information, achieving a significant quality-speed trade-off.

\subsection{Transformer-based Architectures}
Recently, vision transformers have attracted great attention \cite{ViT,SwinT}, and many works have been proposed to explore transformer-based architectures for image restoration \cite{IPT,SwinIR,ELAN,EDT,GRL,HAT}. Various pre-trained models with full or local window-based attention have been exploited and applied to target restoration tasks \cite{IPT,EDT,HAT,HP_transSR}. Liang \textit{et al.} \cite{SwinIR} first introduced the Swin Transformer into image restoration tasks and proposed the hierarchical architecture SwinIR. Cai \textit{et al.} \cite{HP_transSR} proposed the Hierarchical Patch-based Transformer architecture, which significantly enhances single-image super-resolution by progressively recovering high-resolution images through a hierarchy of patch partitions. Recognizing the importance of locality in vision tasks, Li \textit{et al.} proposed LocalViT \cite{LocalViT}, which introduces depth-wise convolution into the feed-forward network of vision transformers. This approach aims to combine the global modeling capabilities of transformers with the local feature extraction strengths of CNNs. LocalViT demonstrated significant performance improvements over baseline models with minimal increases in parameters and computational cost. Several works have focused on lightweight transformer-based models \cite{SwinIR,TCSR,OMNI,NGramSwin}. Wu \textit{et al.} \cite{TCSR} proposed the lightweight model TCSR, which introduces a sliding-window-based self-attention mechanism. One advantage of using transformers for SISR is their ability to capture global context information, beneficial for generating high-quality HR images. However, compared to CNN-based models, transformer-based methods usually require much more computational resources, even with a small model capacity, such as SwinIR-light \cite{SwinIR}.
\subsection{Modern CNN-based Architectures}

Several works have investigated modern CNN-based architectures \cite{convnet,largekernel,SLAK,Conv2Former,ConvViT}. On one hand, large kernel convolutions have been revisited \cite{convnet,largekernel,SLAK}. Liu \textit{et al.} \cite{convnet} explored a modern CNN-based architecture and introduced larger kernels utilizing a $7\times7$ kernel size. Building upon this work, Ding \textit{et al.} \cite{largekernel} increased the kernel size up to 31. Subsequently, Liu \textit{et al.} \cite{SLAK} extended the kernel size up to 51 through sparse training. These advancements in kernel size have shown improvements in capturing complex image details and enhancing super-resolution performance.

On the other hand, many works have focused on hierarchical architectures combining convolutional and transformer elements \cite{Conv2Former,ConvViT}. These architectures leverage the strengths of both convolutional and transformer networks to effectively capture local and global information. Inspired by these findings, we exploit a simple transformer-like ConvNet for lightweight SR tasks, where we replace the self-attention module with a large kernel-based mixer and improve the feed-forward network to preserve more high-frequency information. By leveraging the advantages of large kernel convolutions and transformer-like architectures, the proposed method holds potential for achieving better results in SISR while maintaining computational efficiency.

\subsection{Re-parameterization Methods}

Ding \textit{et al.} \cite{RepVGG} proposed RepVGG, which provides a practical network architecture and the concept of re-parameterization. By re-parameterizing, multiple linear convolutions learned during training can be merged into a single convolution during inference without introducing extra computational costs. For the SISR task, Wang \textit{et al.} \cite{RepSR} proposed RepSR, a plain architecture for SISR via re-parameterization. The authors analyzed the impact of the Batch Normalization (BN) operation in SISR and successfully reintroduced BN into SR. Zhang \textit{et al.} \cite{ECBSR} proposed ECBSR, which introduces more first-order and second-order gradient information into the vanilla convolution via re-parameterization. Wang \textit{et al.} \cite{RepDySR} proposed DDistill-SR, which combines re-parameterization with dynamic convolution \cite{dynamicConv} to extend the learnable landscape while introducing less complexity in inference.

In this study, we introduce the Convolutional Transformer-based Super-Resolution network (CFSR), a simple yet effective model for lightweight image super-resolution. Drawing inspiration from large kernel methods like ConvNet \cite{convnet} and re-parameterization strategies exemplified by RepVGG \cite{RepVGG}, our approach is particularly informed by developments in lightweight super-resolution models such as PAN \cite{PAN} and ECBSR \cite{ECBSR}. CFSR marks a departure from the pixel-attention reliance of PAN, revisiting and streamlining convolution and self-attention feature extraction mechanisms. We introduce an advanced large kernel feature mixer, engineered to deliver exceptional performance and a significantly expanded receptive field. Additionally, our model capitalizes on the successes of transformer-based architectures, integrating a novel edge-enhanced re-parameterization operation into the feed-forward network. This innovation not only enhances local feature extraction but also preserves a greater amount of high-frequency information, a significant advancement over the convolution-centric approach of ECBSR \cite{ECBSR}. By extending the re-parameterization branch and focusing on high-frequency detail retention, CFSR stands at the forefront of the current wave of convolutional technique advancements, uniquely tailoring these approaches for the intricate demands of super-resolution tasks.

\section{Proposed Method \label{sec:method}}
In this paper, we propose a novel lightweight SISR method called CFSR network to leverage large kernel convolutions as gate layers and replace the self-attention module present in transformers. This design enables efficient handling of long-range dependencies and extensive receptive fields while maintaining a lightweight computational cost. Additionally, we introduce the edge-preserving feed-forward network (EFN). EFN incorporates significant image gradient prior, thereby providing more high-frequency information. Furthermore, by re-parameterizing, EFN is free to improve the performance without any extra costs. In this section, we will present the detailed implementation of the propose CFSR.

\subsection{Overall Network Architecture}

In Fig. \ref{fig:first}, we illustrate the proposed CFSR framework, which encompasses three pivotal stages: shallow feature extraction, deep feature extraction, and the image reconstruction module. The shallow feature extraction phase is designed to distill low-level image features, such as edges, textures, and fine-grained details, from the input image and map them into a latent space. These features are important for preserving the local structure and details of the image. In the deep feature extraction stage, the model extracts higher-level, more compact feature representations. It captures  texture and structure information, which is essential for recovering lost details and enhancing the clarity of the image. Utilizing both shallow and deep features, the image reconstruction module is capable of generating high-resolution images. In the following, we will present the details of these three components.

\textbf{Shallow feature extraction.} Given a low-resolution (LR) input image $I_{LR}\in\mathbbm{R}^{H\times W\times 3}$, where $H$ and $W$ are the height and width of this image. The shallow feature extraction utilizes a $3\times 3$ convolution layer to map $I_{LR}$ into the latent feature space,  It can be formulated as:
\begin{equation}
    F_{sf}=\text{H}_{sf}(I_{LR}),
\end{equation}
where $\text{H}_{sf}(\cdot)$ denotes the convolutional layer for shallow feature extraction,  $F_0\in\mathbbm{R}^{H\times W\times C}$ is the output shallow feature, and $C$ is the number of channels.

\textbf{Deep feature extraction.} Then we use a stack of two basic residual blocks (BRB), which contains several ConvFormer layers (CFL), and a $3\times 3$ convolution layer is added at the end of the BRB to aggregate the local features. Specifically, the detailed implementation of BRB and CFL are presented in Fig. \ref{fig:first}, respectively. This process of deep feature extraction is formulated as:
\begin{equation}
    F_k=\text{BRB}_{k}(F_{k-1}),
\end{equation}
where $\text{BRB}_{k}(\cdot)$ denotes the $k$-th BRB. $F_{k-1}$ and $F_{k}$ are the input feature and the output feature of the $k$-th BRB, respectively. Finally, the total deep feature extraction is:
\begin{equation}
    F_{df}=\text{H}_{df}(F_{sf}),
\end{equation}
where $\text{H}_{df}(\cdot)$ presents the general deep feature extraction of the proposed CFSR network, and $F_{df}$ presents the output of the deep backbone.

\textbf{Image reconstruction.} Image reconstruction module aims to reconstruct a high-resolution image $I_{SR}\in \mathbbm{R}^{rH\times rW\times 3}$ by aggregating both shallow and deep features as:
\begin{equation}
    I_{SR}=\text{H}_{rec}(F_{sf}+F_{df}),
\end{equation}
where $r$ is a scale factor. $\text{H}_{rec}(\cdot)$ represents the reconstruction module, which comprises of a $3\times 3$ convolution layer and a pixel-shuffle operation. 

\textbf{Loss function.} The optimization of CFSR parameters is achieved through the minimization of the $L_1$ pixel loss, which can be formulated as:
\begin{equation}
    L_{1}=\left \| I_{SR}-I_{HR}\right \|_{1},
\end{equation}
where $I_{HR}$ is the corresponding ground-truth high-resolution image.

\subsection{ConvFormer Layer}
This section introduces a streamlined, fully convolutional network backbone for lightweight computing, aimed at reducing computational complexity and memory usage.  Firstly, we introduce a large kernel convolution-based feature mixer module, which requires less computational costs while provide effective large receptive fields. Secondly, we introduce the proposed EFN, which induces more high-frequency information for SR model.

\begin{table}[!ht]
\renewcommand\arraystretch{1.2}
    \caption{Comparison of computational costs between global self-attention (SA), local window self-attention (LWSA), and the proposed large kernel mixer (LK), where $K$ is the window/kernel size.    \label{tab:attention_complexity}}
    %\vskip-5pt

    \centering
    \begin{tabular}{ccc}
        \toprule
        \textbf{Module} & \textbf{Complexity} & \textbf{Parameters}\\
        \midrule
        \textbf{SA} & $\mathcal{O}\left(  4 H W C^2 + 2 H^2 W^2 C \right)$ & $4C^2$ \\
        \textbf{LWSA} & $\mathcal{O}\left(  4 H W C^2 + 2 H W C K^2 \right)$ & $4C^2$\\
        \textbf{LK} & $\mathcal{O}\left( 3H W C^2  + H W C K^2 \right)$
        & $3C^2 + CK^2$\\
        \bottomrule
    \end{tabular}
    %\vskip-20pt
\end{table}
\textbf{Large kernel mixer.}
Self-attention is a powerful feature extractor, but its high computational cost makes it impractical for real-time application. Recent studies have demonstrated the effectiveness of employing large kernel convolutions \cite{largekernel}. In Tab. \ref{tab:attention_complexity}, a detailed comparison of the computational complexity between multi-head self-attention (MHSA) \cite{ViT}, local window self-attention (LWSA) \cite{SwinT}, and large kernel convolution (LK) \cite{largekernel} is exhibited.

Here the $H, W, C$, and $K$ represent the height, width, channel dimension, and kernel (local window) size, respectively. The comparison reveals that, given identical window and kernel sizes, the computational complexity of the LK token mixer is significantly less than that of MHSA. The complexity of LK approximates to half that of LWSA, while simultaneously exhibiting a reduction in parameter count, where $K<<C$. Consequently, large kernel convolutions offer a more resource-efficient choice for the design of lightweight models. In this paper, we propose a simple yet effective feature mixer module as  presented in Fig. \ref{fig:first}. Here we take a $1\times1$ convolution followed by a large kernel convolution as the feature mixing gate. Feature extraction of our LK mixer is formulated as follows:
\begin{equation}
\begin{aligned}
    V & = \text{Conv}_{1\times1}(F),\\
    F_{gate} & = \text{DwConv}_{k\times k}(\text{Conv}_{1\times1}(F)),\\
    F_{out} & = \text{Conv}_{1\times1}(V \odot F_{gate}),
\end{aligned}
\end{equation}
where $F$ is the input of the ConvFormer layer, $F_{out}$ is output of it, $\odot$ presents per-pixel production, $k$ is the kernel size, and $DwConv$ is depth-wise convolution.

\begin{figure}[tbp!]
\centering
\includegraphics[width=0.415\textwidth]{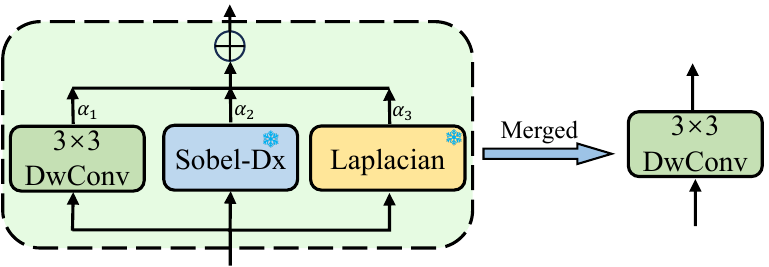}
\vspace{-0.13cm}
\caption{Illustration of the edge-preserving depth-wise convolution (EDC). It contains a multi-branch structure with pre-defined gradient kernels and is equivalent to a single $3\times3$ depth-wise convolution in inference by re-parameterizing.}
\label{fig:edc_kernel}
\end{figure}

\subsection{Edge-preserving Feed-forward Network} 
In the feed-forward network (FFN) in a vanilla Transformer unit, previous studies have enhanced it by integrating depth-wise convolution, thereby improving the local feature ensemble\cite{Restormer,Uformer}. Considering that SISR, being an ill-posed problem, aims at learning an inversion from LR to HR, where the high-frequency information is crucial to it \cite{fang2020soft}. To obtain more high-frequency information in latent features, we propose an edge-preserving feed-forward network (EFN), by our Edge-preserving Depth-Wise Convolution (EDC), as illustrated in Fig. \ref{fig:first}. This allows a best of both words for a richer high-frequency information while maintaining local feature ensemble. The implementation of the proposed EFN is as follows:
\begin{equation}
\begin{aligned}
    F_{1} &= \text{Conv}_{1\times1}(F_{in}),\\
    F_{2} &= \text{GELU}(F_{1}),\\
    F_{EDC} &= \text{EDC}(F_{2}),\\
    F_{3} &= \text{Conv}_{1\times1}(F_{EDC}),
\end{aligned}
\end{equation}
where $\mathrm{GELU}(\cdot)$ is the activation function.

Detailed implementation of our EDC is presented in Fig. \ref{fig:edc_kernel}. It takes a multi-branch structure, containing a standard depth-wise convolution (DwConv) and three DwConvs with pre-defined gradient kernels. Denote $K_{3\times3}\in \mathbb{R}^{C\times1\times3\times3}$ and $B_{3\times3}$ the learnable kernel weights and bias for the vanilla DwConv, where $C$ presents the output channels and 3 is the spatial size. The feature extraction is formulated as:
\begin{equation}
    F_{3\times3} = K_{3\times3}*F_{2}+B_{3\times3},
\end{equation}
where $*$ presents the depth-wise convolution operation.

Next, we take the 1st-order and 2nd-order gradient kernels, such as Sobel filters and Laplacian filters. Denote the $K_{D_x}$, $K_{D_y}$the horizontal and vertical Sobel filters:

\begin{equation}
\small
K_{D_x}=\left[\begin{array}{ccc}
+1 & 0 & -1 \\
+2 & 0 & -2 \\
+1 & 0 & -1
\end{array}\right],~
 K_{D_y}=\left[\begin{array}{ccc}
+1 & +2 & +1 \\
0 & 0 & 0 \\
-1 & -2 & -1
\end{array}\right].
\end{equation}

To align with the shape of kernel $K_{3\times3}$ in depth-wise convolution, we simply expand and repeat the Sobel filters into the $C\times1\times3\times3$ size. The 1st-order gradient of latent feature map is extracted as:
\begin{equation}
    F_{Sobel} =K_{D_x}*F_{2}+B_{D_x}+K_{D_y}*F_{2}+B_{D_y},
\end{equation}
where $B_{D_x} \text{and} ~ B_{D_y}$ are biases.

Moreover, Laplacian filter $K_{Lap}$ is utilized to extract 2nd-order gradient, where we take the 4-neighborhood and 8-neighborhood Laplacian operator as follows:
\begin{equation}
\scriptsize
K_{Lap4}=\left[\begin{array}{ccc}
0 & +1 & 0 \\
+1 & -4 & +1 \\
0 & +1 & 0

\end{array}\right],~K_{Lap8}=\left[\begin{array}{ccc}
+1 & +1 & +1 \\
+1 & -8 & +1 \\
+1 & +1 & +1
\end{array}\right],
\end{equation}
and the same reshaping operation is adopted as aforementioned to extract the 2nd-order intermediate feature:
\begin{equation}
    F_{Lap} = K_{Lap4}*F_{2}+B_{Lap4}+K_{Lap8}*F_{2}+B_{Lap8}.
\end{equation}

The full feature extraction in EDC layer is:
\begin{eqnarray}
    F_{EDC}= \alpha_{1} F_{3\times3}+\alpha_{2}F_{Sobel}+\alpha_{3}F_{Lap},
\end{eqnarray}
where the parameters $\alpha_1,\alpha_2,\alpha_3$  function as learnable competition coefficients for each branch. These coefficients are regulated by a straightforward softmax function, which aids in maintaining a higher retention of high-frequency feature information within the EFN framework.

\textbf{Merged EDC by re-parameterization in inference.} Following \cite{RepVGG,ECBSR}, we can merge the multi-branch EDC layer into one single DwConv in inference without introducing additional complexity. Denote $K$ and $B$ the merged kernel weight and bias of the vanilla DwConv in inference. They can be achieved as follows:
\begin{align}
K&=\alpha_1 K_{3\times3}+\alpha_{2}(K_{D_x}+K_{D_y})+\alpha_{3}(K_{Lap4}+K_{Lap8}), \\
B&=\alpha_1 B_{3\times3}+\alpha_{2}(B_{D_x}+B_{D_y})+\alpha_{3}(B_{Lap4}+K_{Lap8}).
\end{align}

Finally, we merge the five branches into one single DwConv operation by re-parameterizing, and the feature extraction of EDC layer in inference is:
\begin{equation}
    F_{EDC} = K*F_{2}+B.
\end{equation}

\begin{figure}[tbp]
\centering
\includegraphics[width=0.4\textwidth]{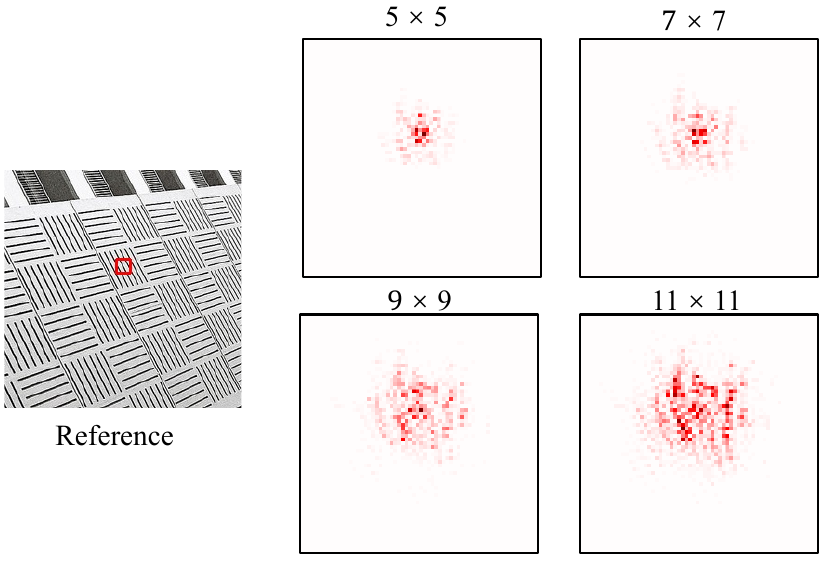}
\vspace{-0.2cm}
\caption{LAM\cite{LAM} attributions of different kernel sizes. From left to right, there is the reference input in the first column and LAM attributions of 5, 7, 9, 11 kernel sizes in the second and third columns.\label{fig:kernel_lam}}
\end{figure}

\begin{table*}[!h]

\caption{\label{tab:result}Quantitative comparison with some state-of-the-art SR approaches on five widely used benchmark datasets. Mult-Adds is evaluated on a $1280 \times 720$ HR image. Results of ours are in \textbf{bold}. }
% One can find that our CFSR achieves SOTA performance among existing advanced CNN-based methods and significantly bridge the gap between Transformer-based methods.

\centering
\renewcommand\arraystretch{1.175}
\resizebox{0.7\textwidth}{!}{
\begin{tabular}{lcllccccc}
\hline
Method                 & \multicolumn{1}{l}{Scale} & \multicolumn{1}{l}{Params} & FLOPs & \begin{tabular}[c]{@{}c@{}}Set5\\ PSNR/SSIM\end{tabular} & \begin{tabular}[c]{@{}c@{}}Set14\\ PSNR/SSIM\end{tabular} & \begin{tabular}[c]{@{}c@{}}BSD100\\ PSNR/SSIM\end{tabular} & \begin{tabular}[c]{@{}c@{}}Urban100\\ PSNR/SSIM\end{tabular} & \begin{tabular}[c]{@{}c@{}}Manga109\\ PSNR/SSIM\end{tabular} \\ 
\hline
\hline

IMDN  \cite{IMDN}&                           & 694K                       & 158.8G    & 38.00/0.9605                                             & 33.63/0.9177                                              & 32.19/0.8996                                               & 32.17/0.9283                                                & 38.88/0.9774                                                \\

% LatticeNet\cite{LatticeNet}& & 756K & 169.5G & 38.15/0.9610 & 33.78/0.9193 & 32.25/0.9005 & 32.43/0.9302 & -/- \\

LAPAR-A\cite{LAPAR} &     & 548K  & 171.0G   & 38.01/0.9605  & 33.62/0.9183 & 32.19/0.8999 & 32.10/0.9283 & 38.67/0.9772\\

% SMSR\cite{SMSR} & & 985K  & 131.6G   & 38.00/0.9601 & 33.64/0.9179  & 32.17/0.8990 & 32.19/0.9284   & 38.76/0.9771 \\
ECBSR\cite{ECBSR} & & 596K & 137.3G & 37.90/0.9615 & 33.34/0.9178 &   32.10/{0.9018} & 31.71/0.9250 & -/- \\

PAN \cite{PAN}& & 261K & 70.5G     & 38.00/0.9605   & 33.59/0.9181  & 32.18/0.8997 & 32.01/0.9273  & 38.70/0.9773   \\
DRSAN\cite{DRSAN} && 370K & 85.5G & 37.99/{0.9606} & 33.57/0.9177 & 32.16/0.8999 & 32.10/0.9279 & -/- \\

% DDistill-SR\cite{DDistill} & &414K & 128.0G&38.03/{0.9606}& 33.61/0.9182 & 32.19/0.9000 &32.18/0.9286 &{38.94}/{0.9777} \\
RFDN\cite{RFDN} &    $\times$2                       & 534K                       & 95.0G     & {38.05}/{0.9606}                                             & {33.68}/{0.9184}                                              & 32.16/0.8994                                               & 32.12/0.9278                                                 & 38.88/0.9773                                                 \\

RLFN \cite{Ko_2022_CVPR} &  & 527K  & 116G  & 38.07/0.9607 & 33.72/0.9187 & 32.22/0.9000 & 32.33/0.9299  & -/- \\
DIPNet \cite{yu2023dipnet}  &  & 527K  & 119G  & 37.98/0.9605  & 33.66/0.9192  & 32.20/0.9002  & 32.31/0.9302   & 38.62/0.9770\\

ShuffleMixer\cite{shufflemixer} & & 394K& 91.0G &38.01/{0.9606}& 33.63/0.9180 &32.17/0.8995& 31.89/0.9257& 38.83/0.9774 \\

SPAN \cite{wan2024swift} & & 481K & 94.4G & 38.08/0.9608 &  33.71/0.9183 & 32.22/0.9002 & 32.24/0.9294 &  38.94/0.9777 \\

\textbf{CFSR (Ours)} &                           & 291K                       & 62.6G     & {\textbf{38.07}}/{\textbf{0.9607}}                   & {\textbf{33.74}}/{\textbf{0.9192}}                            & {\textbf{32.24}}/{\textbf{0.9005}}                  & {\textbf{32.28}}/{\textbf{0.9300}}                     & {\textbf{39.00}}/{\textbf{0.9778}}     \\
\hline
SwinIR-light \cite{SwinIR} & & 878K & 195.6G & {38.14}/{0.9611} & {33.86/0.9206} & {32.31/0.9012} & {32.76/0.9340} & {39.12/0.9783} \\
DLGSANet \cite{DLGSANet}&    & {566K}       & {128.1G}    & 38.16/{0.9611} & {33.92}/0.9202 & 32.26/0.9007 & {32.82/0.9343} & {39.14}/0.9777 \\
Omni-SR \cite{omni_sr}& & 772K & {172.1G} & {38.22}/{0.9613} & {33.98}/{0.9210} &  {32.36}/{0.9020} & {33.05}/{0.9363}  & {39.28}/{0.9784} \\

% ELAN-light\cite{ELAN}&  & 582K & 168.4G & {38.17/0.9611} & {33.94/0.9207} & 
% {32.30}/{0.9012} & {32.76/0.9340} & 39.11/{0.9782}\\
\hline
\hline
% VDSR \cite{VDSR}    &  & 666K &612.6G   & 33.66/0.9213 & 29.77/0.8314 & 28.82/0.7976 & 27.14/0.8279 & 32.01/0.9340 \\
% LapSRN \cite{LapSRN} &   & 502K  &90.0G  & 33.81/0.9220 & 29.79/0.8325 & 28.82/0.7980 & 27.07/0.8275 & 32.21/0.9350 \\
% CARN\cite{CARN}& & 1,592K &118.8G & 34.29/0.9255 & 30.29/0.8407 & 29.06/0.8034 & 28.06/0.8493 & 33.50/0.9440 \\
% SRResNet\cite{SRGAN} &  & 1,554K  &190.2G & {34.41}/0.9274 & 30.36/{0.8427} & 29.11/0.8055 & 28.20/0.8535 & 33.54/0.9448 \\

IMDN \cite{IMDN} &    & 703K  &72.0G  & 34.36/0.9270 & 30.32/0.8417 & 29.09/0.8046 & 28.17/0.8519 & 33.61/0.9445 \\
LatticeNet\cite{LatticeNet}& & 765K & 76.3G & 34.40/0.9272 & 30.32/0.8416 & 29.10/0.8049 & 28.19/0.8513 & -/- \\

LAPAR-A \cite{LAPAR}  &     & 594K  & 114.0G  & 34.36/0.9267 & 30.34/0.8421 & 29.11/0.8054 & 28.15/0.8523 & 33.51/0.9441 \\

% SMSR\cite{SMSR}  & & 993K  & 67.8G & 34.40/0.9270 & 30.33/0.8412 & 29.10/0.8050   & {28.25}/{0.8536} & 33.68/0.9445  \\
PAN \cite{PAN}   &  & 261K  & 39.0G & 34.40/0.9271 & 30.36/0.8423 & {29.11}/0.8050 & 28.11/0.8511 & 33.61/0.9448 \\ 

DRSAN\cite{DRSAN} &$\times3$& 410K  & 43.2G & {34.41}/0.9272 & 30.27/0.8413 & 29.08/{0.8056} & 28.19/0.8529 & -/- \\
Distill-SR\cite{DDistill} & & 414K & 57.4G &34.37/{0.9275} &30.34/0.8420  & 29.11/0.8053 &28.19/0.8528 & {33.69}/{0.9451} \\
ShuffleMixer \cite{shufflemixer} & &  415K & 43.0G &34.40/0.9272 & {30.37}/0.8423 & {29.12}/0.8051 & 28.08/0.8498 & {33.69}/0.9448  \\

\textbf{CFSR (Ours)} &  &   298K& 28.5G & {\textbf{34.50}}/{\textbf{0.9279}} &  {\textbf{30.44}}/{\textbf{0.8437}} & {\textbf{29.16}}/{\textbf{0.8066}} & {\textbf{28.29}}/{\textbf{0.8553}}  & {\textbf{33.86}}/{\textbf{0.9462}} \\ 
\hline
SwinIR-light \cite{SwinIR} & & 886K & 87.2G  & {34.62/0.9289} & {30.54}/{0.8463} & {29.20}/{0.8082} & {28.66}/{0.8624} & {33.98}/{0.9478}\\
DLGSANet \cite{DLGSANet}&   & {572K}       & {56.8G}     &{34.63}/0.9288 &{30.57}/0.8459 &{29.21/0.8083} &{28.69/0.8630} &{34.10/0.9480} \\
Omni-SR\cite{omni_sr}  &  & 780K & {78.0G}  & {34.70}/{0.9294} & {30.57}/{0.8469} & {29.28}/{0.8094} & {28.84}/{0.8656} & {34.22}/{0.9487} \\

% ELAN-light \cite{ELAN}& & 590K & 75.7G &  {34.61/0.9288} & {30.55/0.8463} & {29.21}/{0.8081} & {28.69/0.8624} & {34.00/0.9478}\\
             \hline
             \hline

IMDN\cite{IMDN} &                           & 715K                       & 40.9G     & 32.21/0.8948                                             & 28.58/0.7811                                              & 27.56/0.7353                                               & 26.04/0.7838                                                 & 30.45/0.9075                                                 \\
LatticeNet\cite{LatticeNet}& & 777K & 43.6G & 32.18/0.8943 & 28.61/0.7812 & 27.57/0.7355 & 26.14/0.7844 & -/- \\

LAPAR-A \cite{LAPAR}  &                       & 659K                       & 94.0G                         & 32.15/0.8944                                                                 & 28.61/0.7818                                                                  & 27.61/{0.7366}                                                                 & 26.14/0.7871                                                                     & 30.42/0.9074                                                                     \\
% SMSR\cite{SMSR}   &                        & 1006K                      & 41.6G                         & 32.12/0.8932                                                                 & 28.55/0.7808                                                                  & 27.55/0.7351                                                                   & 26.11/0.7868                                                                     & 30.54/0.9085                                                                     \\
ECBSR\cite{ECBSR}   & & 603K& 34.7G& 31.92/0.8946& 28.34/0.7817 &27.48/0.7393 &25.81/0.7773& -/- \\
PAN\cite{PAN} &                           & 272K                       & 28.2G     & 32.13/0.8948                                             & {28.61}/{0.7822}                                              & {27.59}/0.7363                                               & {26.11}/0.7854                                                 & 30.51/{0.9095}                                                 \\

DRSAN\cite{DRSAN} & &410K&30.5G & 32.15/0.8935 & 28.54/0.7813 & 27.54/0.7364 & 26.06/0.7858 & -/- \\

DDistill-SR\cite{DDistill} &$\times$4  & 434K &33.0G &32.23/{0.8960} &{28.62}/0.7823 & 27.58/0.7365 &  {26.20}/{0.7891} & 30.48/0.9090 \\

RFDN  \cite{RFDN}&                           & 550K                       & 23.9G     & {32.24}/{0.8952}                                             & {28.61}/0.7819                                              & 27.57/0.7360                                               & {26.11}/{0.7858}                                                 &{30.58}/0.9089                                                 \\
RLFN  \cite{Kong2022ResidualLF} &  & 543K & 29.8G & 32.24/0.8952 & 28.62/0.7813 & 27.60/0.7364 & 26.17/0.7877 & -/- \\
DIPNet  \cite{yu2023dipnet} &  & 543K & 30.9G & 32.20/0.8950 & 28.58/0.7811 & 27.59/0.7364 & 26.16/0.7879 & 30.53/0.9087 \\
ShuffleMixer\cite{shufflemixer} & & 411K &28.0G& 32.21/0.8953 &28.66/{0.7827} &{27.61}/{0.7366} & 26.08/0.7835 &{30.65}/0.9093 \\
SPAN \cite{wan2024swift} & &  498K & 24.5G & 32.20/0.8953 & 28.66/0.7834 & 27.62/0.7374 & 26.18/0.7879 & 30.66/0.9103 \\
\textbf{CFSR (Ours)} & & 307K                       & 17.5G     & {\textbf{32.33}}/{\textbf{0.8964}}                                             & {\textbf{28.73}}/{\textbf{0.7842}}                                              & {\textbf{27.63}}/{\textbf{0.7381}}                                               & {\textbf{26.21}}/{\textbf{0.7897}}                                                 & {\textbf{30.72}}/{\textbf{0.9111}}    \\ 
\hline
SwinIR-light \cite{SwinIR} && 897K & 49.6G & {32.44/0.8976} & {28.77}/{0.7858} & {27.69/0.7406} & {26.47/0.7980} & {30.92/0.9151}\\
DLGSANet \cite{DLGSANet} & & {581K} & {32.0G}     & {32.46/0.8984} & {28.79/0.7861} & {27.70/0.7408} & {26.55/0.8002} & {30.98}/0.9137 \\
Omni-SR \cite{omni_sr} &  & 792K & {45.0G} & {32.49}/{0.8988} & {28.78}/{0.7859}  & {27.71}/{0.7415}  & {26.64}/{0.8018} &  {31.02}/{0.9151} \\

% ELAN-light\cite{ELAN} & & 601K & 43.2G & {32.43/0.8975} & {28.78/0.7858} & {27.69/0.7406} & {26.54/0.7982} & {30.92}/{0.9150}\\

\hline
\end{tabular}
}
\vspace{0.4cm}
\end{table*}

\section{EXPERIMENTS AND ANALYSIS \label{sec:experiments}}
In this section we will describe the detailed evaluation experiments. Firstly, we introduce the experiment settings and comparison methods. Then quantitative and qualitative results are reported on some public datasets of SOTA light-weight methods and our proposed method. Lastly, to verify the technical contribution of the proposed method, we present the performance of different variants of the proposed method through some ablation studies.

\begin{figure*}[!htbp]
\centering
\includegraphics[width=0.85\textwidth]{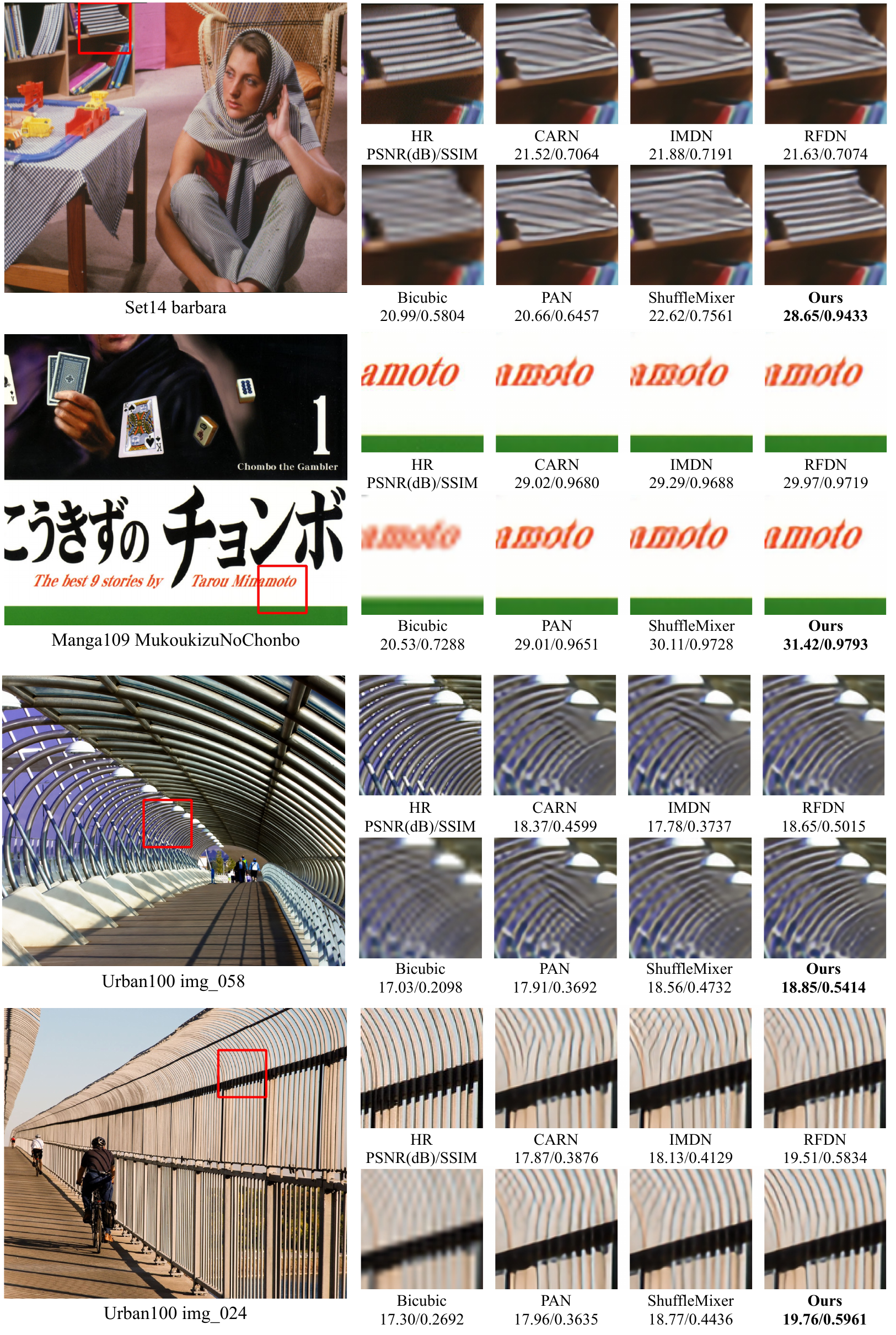}
\vskip-10.pt
\caption{Visual comparisons for SR($\times$4) methods on Set14, Manga109, and Urban100 datasets (\textbf{Zoom in for more details}).}
\label{fig:visual}
\end{figure*}

\begin{figure*}[htbp]
\centering
\includegraphics[width=0.9\textwidth]{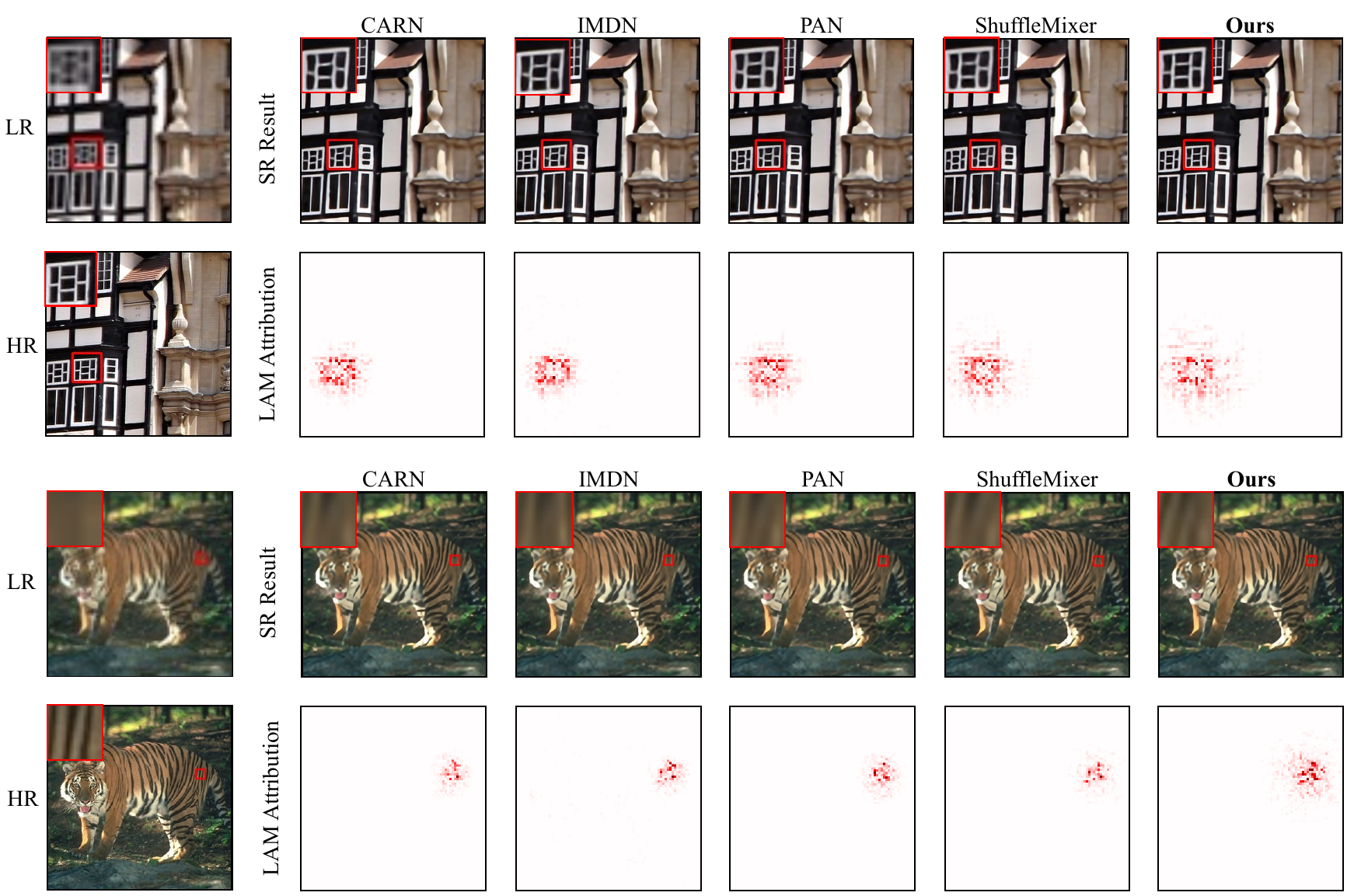}
\vspace{-0.3cm}
\caption{LAM\cite{LAM} comparisons between SOTA methods and the proposed CFSR. Results of two samples from Urban100 and B100 datasets are presented. One can find that the proposed CFSR outperforms other advanced models with larger receptive fields and richer textures (\textbf{Zoom in for more details.}). }
\label{fig:compare_lam}
\end{figure*}

\subsection{Experimental Setup}
\textbf{Datasets and evaluation metrics.} Following comparison methods\cite{LAPAR,PAN,DDistill,shufflemixer}, we train our model on the DIV2K \cite{DIV2K} and Flickr2K \cite{EDSR} datasets, which contain 3450 high-quality images. We test the performance of CFSR on five benchmark test datasets, including Set5 \cite{Set5}, Set14 \cite{Set14}, BSD100 \cite{B100}, Urban100 \cite{Urban100} and Manga109 \cite{Manga109}. We evaluate the Peak Signal-to-Noise Ratio (PSNR) and Structural Similarity Index Measure (SSIM) on Y channel of transformed YCbCr space. 

\textbf{Training details.} The channel number, RCFB number and CFL number in each RCFB are set to 48, 2 and 6, respectively. The sizes $k_1$ and $k_2$ of two depth-wise convolution in CFL are set to 9 and 3. During training, we randomly crop the image patches with the fixed size of 64 × 64 and set the batch size to 16 for training. We employ randomly rotating $90^{\circ}$, $180^{\circ}$, $270^{\circ}$ and horizontal flip for data augmentation. We use ADAM \cite{ADAM} with $\beta_1=0.9$ and $\beta_2=0.99$ to optimize $L_1$ loss. The initial learning rate is 2e-4. The CFSR is implemented by PyTorch \cite{Pytorch} and trained with Nvidia RTX A4000 GPU.

\textbf{Comparison Methods}
We compare the proposed CFSR with state-of-the-art efficient SR approaches, including IMDN \cite{IMDN}, LatticeNet\cite{LatticeNet}, LAPAR\cite{LAPAR}, SMSR\cite{SMSR}, ECBSR\cite{ECBSR}, DRSAN\cite{DRSAN}, PAN \cite{PAN}, DDistill\cite{DDistill}, RFDN \cite{RFDN}, RLFN \cite{Ko_2022_CVPR}, ShuffleMixer\cite{shufflemixer}, DIPNet \cite{yu2023dipnet}, SPAN \cite{wan2024swift} and some recent Transformer-based methods, including SwinIR-light \cite{SwinIR}, DLGSANet \cite{DLGSANet} and Omni-SR \cite{omni_sr}.

\subsection{Main Results}
The proposed CFSR achieves promising performance with less model complexity in both quantitative and qualitative results.

\textbf{Quantitative evaluation.}
Table \ref{tab:result} presents quantitative comparisons for the upscaling factors of 2x, 3x, and 4x on five test datasets, including parameter counts and FLOPs for each method. Remarkably, our proposed CFSR outperforms existing advanced CNN-based methods in terms of both PSNR and SSIM across all scales and datasets, and significantly bridges the gap between Transformer-based methods\cite{SwinIR,DLGSANet,omni_sr}.
When compared to DDistill-SR \cite{DDistill}, CFSR maintains superior performance across all scales and datasets, while exhibiting approximately half the computational complexity. For 3x super-resolution tasks, CFSR attains significant performance gains over ShuffleMixer \cite{shufflemixer}, enhancing PSNR by \textbf{0.19} dB and \textbf{0.17} dB, and SSIM by \textbf{0.0055} and \textbf{0.0014} on the Urban100 and Manga109 test datasets, respectively.
Notably, CFSR demonstrates competitive performance against recent NTIRE challenge winners, including RLFN \cite{Kong2022ResidualLF}, DIPNet \cite{yu2023dipnet}, and SPAN \cite{wan2024swift}. Across these comparisons, CFSR consistently achieves comparable or superior performance in terms of PSNR and SSIM.

These results underscore CFSR's potential as a resource-efficient and performance-oriented model for lightweight image super-resolution tasks. By delivering competitive performance with significantly reduced computational demands, CFSR represents a notable advancement in balancing efficiency and effectiveness in the field of image super-resolution.

\begin{table}
\renewcommand\arraystretch{1.2}
\centering
\caption{\label{tab:kernel}Ablation on the size of large kernel convolution for $\times4$ SR. We test the results on Urban100 and Manga109 datasets.}
\begin{tabular}{ccccc}
\toprule
Kernel Size & \multicolumn{1}{l}{Params} & FLOPs & \begin{tabular}[c]{@{}c@{}}Urban100\\ PSNR/SSIM\end{tabular} & \begin{tabular}[c]{@{}c@{}}Manga109\\ PSNR/SSIM\end{tabular} \\ 
\midrule
$5\times5$         & 274K                       & 15.6G     & 26.02/0.7834                                                 & 30.47/0.9082                                                 \\
$7\times7$         & 288K                       & 16.4G     & 26.08/0.7854                                                 & 30.54/0.9089                                                 \\
$9\times9$         & 307K                       & 17.5G     & 26.13/0.7875                                                 & 30.60/0.9098                                                 \\ 
$11\times11$       & 330K                       & 18.9G     & 26.16/0.7880                                                 & 30.64/0.9102                                                 \\
\bottomrule
\end{tabular}
\end{table}

\textbf{Qualitative evaluation.}
We conducted a visual quality comparison of SR results between our proposed CFSR and five representative models, including CARN \cite{CARN}, IMDN \cite{IMDN}, RFDN \cite{RFDN}, PAN \cite{PAN}, and ShuffleMixer \cite{shufflemixer}. The $\times4$ SR results are presented in Fig. \ref{fig:visual}. When we take a closer look at the results in the second column, one can find that our CFSR is able to recover the main structures with sharp textures. Moreover, results of samples 'img\_058' and 'img\_024' in Urban100 dataset showcase that CFSR can obtain clearer edges while others fail.

Furthermore, we use LAM\cite{LAM}, which represents the range of attribution pixels, to visualize receptive fields. Visual results are presented in Fig. \ref{fig:compare_lam}, showing that our CFSR can take advantage of a wider range of information than CARN, IMDN, PAN, ShuffleMixer, obtaining large receptive fields effectively. Let us take the 'Tiger' image in the second row of Fig. \ref{fig:compare_lam} as the example. One can find that the proposed CFSR can achieve richer textures with the larger receptive field compared to other advanced methods.

\textbf{Comparison on unknown degradation.}
Given that the primary objective of image Super-Resolution (SR) is to address complex real-world degradations and generate visually appealing images, we conduct comprehensive evaluation on the RealSR \cite{RealSRdata} and unknown DIV2K datasets, known for their intricate degradation patterns.  For a fair comparison, we retrain and evaluate the ShuffleMixer model and our CFSR on these datasets, respectively. The results, as detailed in Fig. \ref{real_result}, clearly demonstrate that our CFSR outperforms the advanced ShuffleMixer in handling complex degradation tasks. 

\begin{figure}[tbp]
\centering
\includegraphics[width=0.475\textwidth]{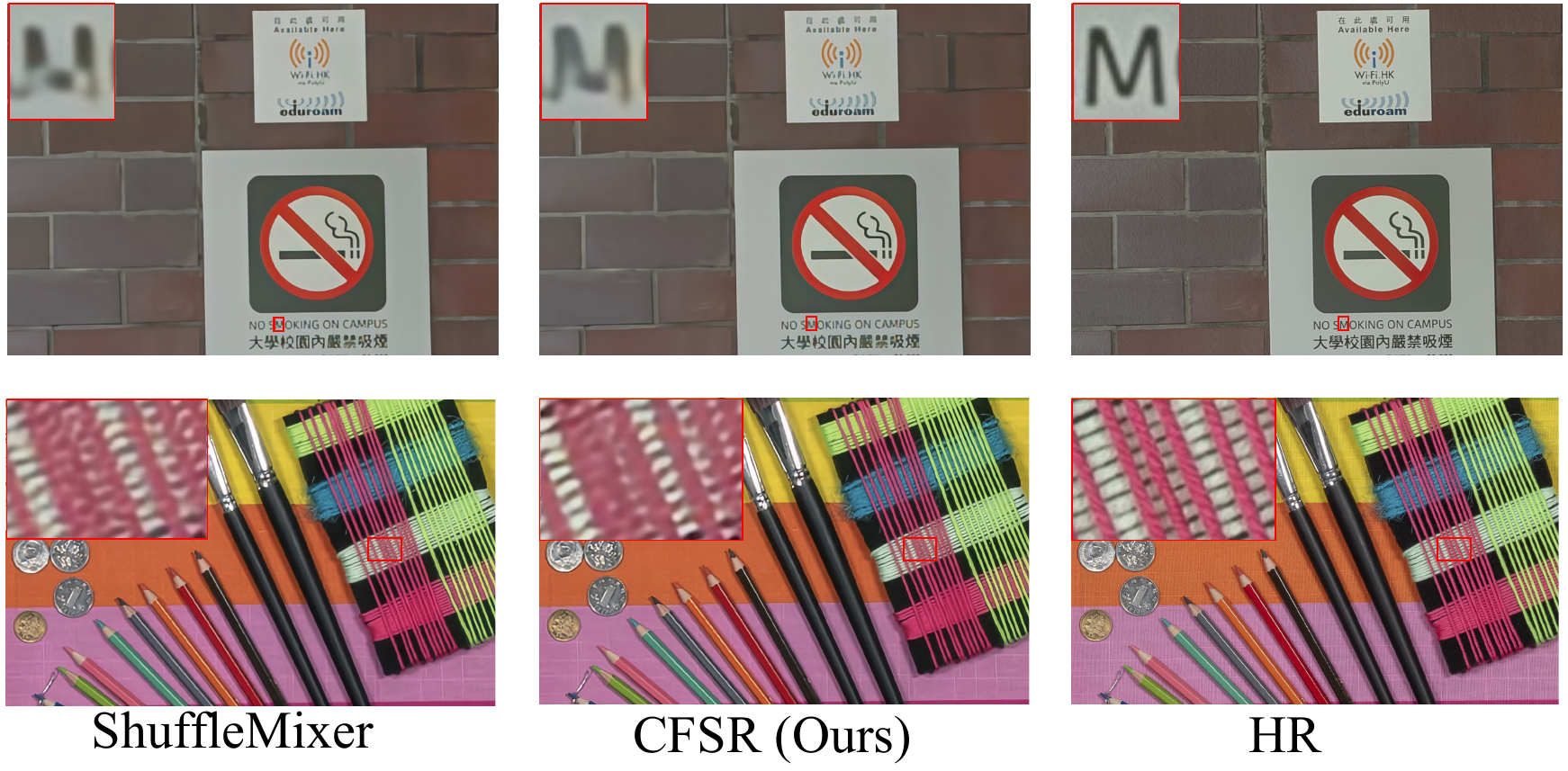}
\vspace{-0.3cm}
\caption{Visual comparison on RealSR dataset. Super-resolved results of our CFSR can achieve more accurate textures even with complex degradation. \label{fig:realsr_demo}}

\end{figure}

Furthermore, in subjective image quality assessments, the CFSR model exhibits superior performance, and some results are presented in Fig. \ref{fig:realsr_demo}. One can find that our CFSR is able to accurately reconstruct finer details and maintain a high level of clarity, even in areas of intricate patterns and textures. We believe these comparisons showcase the practical effectiveness of the CFSR model in real-world applications.

\subsection{Ablation Studies}
In this section, we conduct in-depth ablation studies on the core component of CFSR, the ConvFormer layer. The ConvFormer layer consists of two main elements: the Large Kernel Feature Mixer (LK Mixer) and the Edge-preserving Feed-forward Network (EFN). Each of these is ablated separately to elucidate their individual impact on the overall model performance. Specifically, we analyze the influence of various kernel sizes in the LK Mixer and examine the effect of our proposed Edge-preserving Depthwise Convolution (EDC) within the EFN.

\begin{figure}[!t]
\centering
\includegraphics[width=0.475\textwidth]{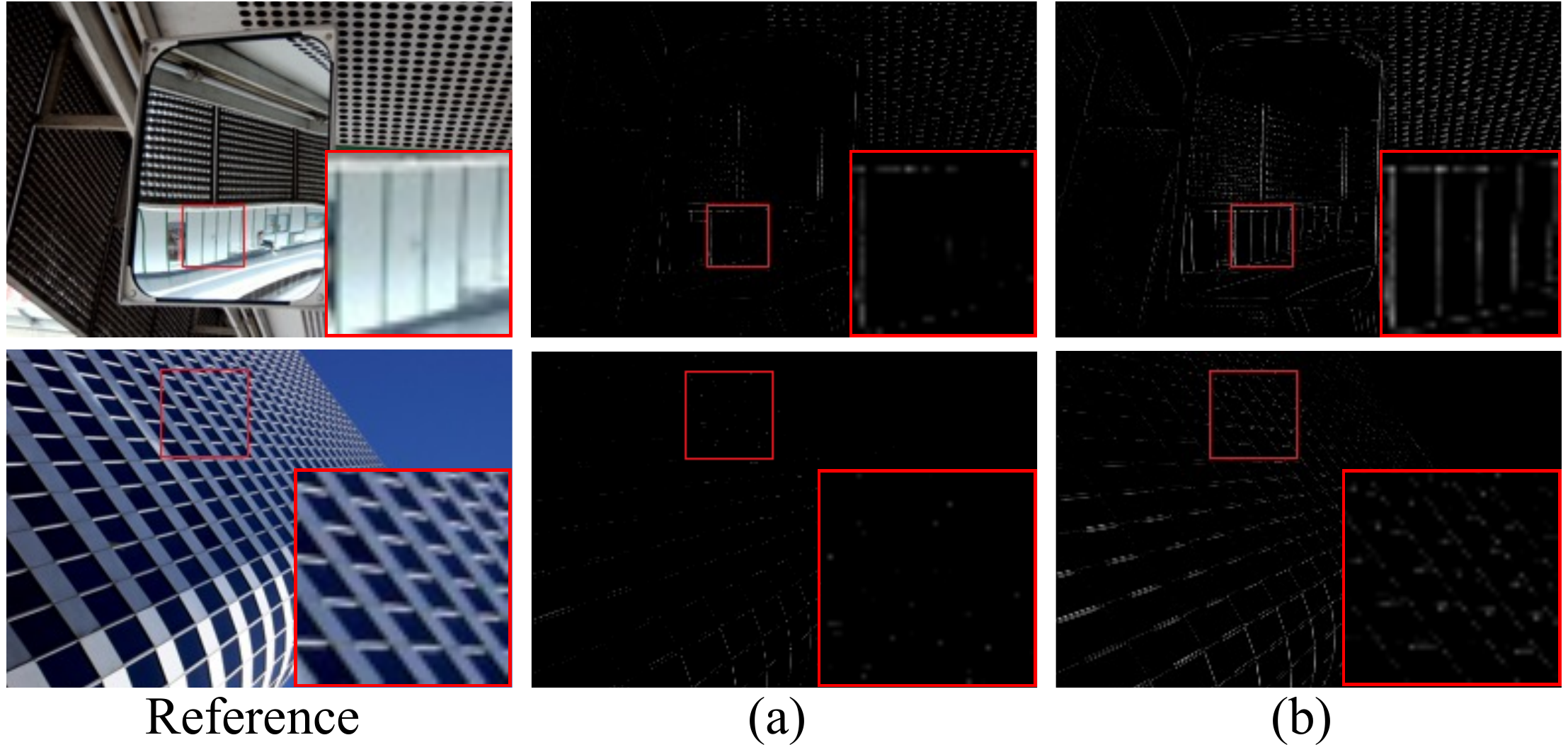}
\vspace{-0.3cm}
\caption{Visual comparisons between latent features learned with or without the EDC (correspond to the ablation results in Tab. \ref{tab:edc_table}). The first column is the reference image sampled from Urban100 dataset. (a) Latent features learned by the vanilla DwConv. (b) Feature maps extracted by our EFN with the EDC layer. One can find that EFN can substantially obtains clear and robust high-frequency information in latent feature maps. }
\label{fig:edc_feature}
\end{figure}

\begin{table}[t]
\renewcommand\arraystretch{1.0}
\centering
\caption{Comparison on unknown degradation. We evaluate the proposed CFSR and ShuffleMixer\cite{shufflemixer} on RealSR test dataset and unknown DIV2K evaluation dataset.} \label{real_result}
\begin{tabular}{lcccc}
\toprule
Method &   Params. & FLOPs & RealSR & DIV2K unknown  \\
\midrule
ShuffleMixer & 411K & 28.0G & 29.16/0.8261 & 29.17/0.8049 \\
\textbf{CFSR} & \textbf{307K} & \textbf{17.5G} & \textbf{29.25}/\textbf{0.8266} & \textbf{29.39}/\textbf{0.8111}\\
\bottomrule
\end{tabular}
\end{table}

\begin{table}[ht]
\renewcommand\arraystretch{1.2}
\centering
\caption{Comparison of different re-parameterized convolution configurations in the EFN for $\times4$ SR. We report the results on Urban100 and Manga109 datasets. Configurations: (a) vanilla depth-wise convolution, (b) repConv with one more depth-wise convolution (c) asymmetric convolution from ACNet, and (d) our proposed EDC.}
\label{tab:edc_table}
\scalebox{0.98}{
\begin{tabular}{ccccc}
\toprule
Method  & \multicolumn{1}{l}{Params} & FLOPs & \begin{tabular}[c]{@{}c@{}}Urban100\\ PSNR/SSIM\end{tabular} & \begin{tabular}[c]{@{}c@{}}Manga109\\ PSNR/SSIM\end{tabular} \\ 
\midrule
%FFN & 295K & 16.9G  & 26.05/0.7851 & 30.50/0.9085  \\
(a) & 307K & 17.5G & 26.13/0.7875  & 30.60/0.9098 \\ 
(b) & 307K & 17.5G & 26.14/0.7875  & 30.62/0.9100 \\ 
(c) & 307K & 17.5G & 26.14/0.7876  & 30.62/0.9099 \\ 
(d) & 307K & 17.5G & \textbf{26.21}/\textbf{0.7897}  & \textbf{30.72}/\textbf{0.9111} \\ 
\bottomrule
\end{tabular}
}
\end{table}

\textbf{Impact of the kernel size.} Our in-depth analysis, as presented in Tab. \ref{tab:kernel}, explores the implications of varying kernel sizes on model performance. The results clearly demonstrate that model performance improves as the kernel size increases. To provide an intuitive understanding, we also present visualizations of the activated pixels using LAM \cite{LAM} in Fig. \ref{fig:kernel_lam}, showcasing the superiority of larger kernels in effectively extending the receptive field.

Examining the results in Tab. \ref{tab:kernel} more closely, we observe that while larger kernel sizes yield superior performance, they also result in increased computational demands. Therefore, we limited our investigation to kernel sizes not exceeding 11 due to both computational constraints and an evident saturation effect. Specifically, a progressive enhancement in PSNR/SSIM values of over 0.05 dB/0.001 was recorded as the kernel size increased from 5 to 9 across both the Urban100 and Manga109 datasets. This consistent gain emphasizes the potency of large kernels in improving performance. However, increasing the kernel size from 9 to 11 resulted in a more modest gain of only 0.03 dB/0.0005 in PSNR/SSIM. Given this negligible improvement and the saturation observed in benefits with larger kernel sizes, we selected 9 as the default kernel size for CFSR. Furthermore, the LAM results for different kernel sizes in the LK Mixer are presented in Fig. \ref{fig:kernel_lam}, illustrating that the proposed LK Mixer in CFSR can effectively provide large receptive fields.

\textbf{Impact of EDC.} 
To thoroughly evaluate the efficacy of our proposed Edge-preserving Depth-wise Convolution (EDC) within the Edge-preserving Feed-forward Network (EFN), we conducted comprehensive experiments comparing various re-parameterized convolution configurations. The results are presented in Tab. \ref{tab:edc_table}. We examined four configurations: (a) vanilla depth-wise convolution, (b)repConv with one more depthwise convolution (c) asymmetric convolution, and (d) our proposed EDC. It is important to note that the Edge-oriented Convolution Block (ECB) was originally designed for dense convolution, which introduces significantly more parameters when adapted to our model structure. Our proposed EDC demonstrates substantial performance improvements compared to the vanilla depth-wise convolution (a) and re-parameterization modules without edge priors (b) and (c). These results highlight the superiority of our EDC's ability in enhancing edge features for super-resolution tasks. The effectiveness of the proposed EFN in leveraging edge extraction priors within the FFN structure is particularly noteworthy. By introducing edge-preserving capabilities, we achieve significant performance gains without incurring additional computational overhead during inference. This balance of improved performance and maintained efficiency is crucial for lightweight super-resolution models.

To provide a more comprehensive understanding, we visualize the latent feature maps in Fig. \ref{fig:edc_feature}. The incorporation of EDC yields a pronounced improvement in the extraction of high-frequency information within intermediate features. Compared to the vanilla depth-wise convolution in the FFN, Figure \ref{fig:edc_feature} clearly shows more distinct edges and textures in the latent feature maps of our EFN. This visual evidence further underscores the substantial benefits gained from integrating edge-preserving priors into the FFN through our re-parameterization approach, highlighting the unique contribution of our method in the context of efficient super-resolution.

\section{Conclusion\label{sec:conclusion}}
In recent years, the field of Single-Image Super-Resolution (SISR) has witnessed significant advancements, largely due to the adoption of deep learning techniques. In this paper, we have proposed a transformer-like convolutional network for lightweight super-resolution tasks, named CFSR, which achieves state-of-the-art performance. CFSR utilizes a large kernel feature mixer (LK Mixer) as an efficient alternative to the computationally intensive self-attention module, effectively modeling extensive receptive fields while substantially reducing computational overhead. Additionally, we have introduced an edge-preserving feed-forward network (EFN) to extract local features while preserving high-frequency information. To further enhance the edge-preserving capabilities of our network, we proposed the edge-preserving depth-wise convolution (EDC), which enriches high-frequency information without adding extra computational complexity during inference through a re-parameterization strategy. We conducted detailed ablation studies to understand the influence of these components. Comprehensive experiments demonstrate the superior performance of CFSR, showcasing its ability to surpass existing state-of-the-art methods while maintaining a lean profile in terms of parameter count and computational complexity. Overall, our proposed method holds significant potential for advancing the field of SISR and facilitating the deployment of super-resolution algorithms on resource-constrained devices.

\bibliographystyle{unsrt}
\bibliography{reference}

\begin{thebibliography}{10}

\bibitem{survey_TMM}
Wenming Yang, Xuechen Zhang, Yapeng Tian, Wei Wang, Jing{-}Hao Xue, and Qingmin Liao.
\newblock Deep learning for single image super-resolution: {A} brief review.
\newblock {\em {IEEE} Trans. Multim.}, 21(12):3106--3121, 2019.

\bibitem{21survey}
Juncheng Li, Zehua Pei, and Tieyong Zeng.
\newblock From beginner to master: {A} survey for deep learning-based single-image super-resolution.
\newblock {\em CoRR}, abs/2109.14335, 2021.

\bibitem{CNN-survey}
Zhihao Wang, Jian Chen, and Steven C.~H. Hoi.
\newblock Deep learning for image super-resolution: {A} survey.
\newblock {\em {IEEE} Trans. Pattern Anal. Mach. Intell.}, 43(10):3365--3387, 2021.

\bibitem{SRCNN}
Chao Dong, Chen~Change Loy, Kaiming He, and Xiaoou Tang.
\newblock Image super-resolution using deep convolutional networks.
\newblock {\em {IEEE} Trans. Pattern Anal. Mach. Intell.}, 38(2):295--307, 2016.

\bibitem{VDSR}
Jiwon Kim, Jung~Kwon Lee, and Kyoung~Mu Lee.
\newblock Accurate image super-resolution using very deep convolutional networks.
\newblock In {\em {CVPR}}, pages 1646--1654, 2016.

\bibitem{EDSR}
Bee Lim, Sanghyun Son, Heewon Kim, Seungjun Nah, and Kyoung~Mu Lee.
\newblock Enhanced deep residual networks for single image super-resolution.
\newblock In {\em CVPRW}, pages 1132--1140, 2017.

\bibitem{RCAN}
Yulun Zhang, Kunpeng Li, Kai Li, Lichen Wang, Bineng Zhong, and Yun Fu.
\newblock Image super-resolution using very deep residual channel attention networks.
\newblock In {\em ECCV}, pages 294--310, 2018.

\bibitem{SAN}
Tao Dai, Jianrui Cai, Yongbing Zhang, Shu{-}Tao Xia, and Lei Zhang.
\newblock Second-order attention network for single image super-resolution.
\newblock In {\em {CVPR}}, pages 11065--11074, 2019.

\bibitem{HAN}
Ben Niu, Weilei Wen, Wenqi Ren, Xiangde Zhang, Lianping Yang, Shuzhen Wang, Kaihao Zhang, Xiaochun Cao, and Haifeng Shen.
\newblock Single image super-resolution via a holistic attention network.
\newblock In {\em {ECCV}}, pages 191--207, 2020.

\bibitem{NLSN}
Yiqun Mei, Yuchen Fan, and Yuqian Zhou.
\newblock Image super-resolution with non-local sparse attention.
\newblock In {\em {CVPR}}, pages 3517--3526, 2021.

\bibitem{FSRCNN}
Chao Dong, Chen~Change Loy, and Xiaoou Tang.
\newblock Accelerating the super-resolution convolutional neural network.
\newblock In {\em {ECCV}}, pages 391--407, 2016.

\bibitem{CARN}
Namhyuk Ahn, Byungkon Kang, and Kyung{-}Ah Sohn.
\newblock Fast, accurate, and lightweight super-resolution with cascading residual network.
\newblock In {\em {ECCV}}, pages 256--272, 2018.

\bibitem{IMDN}
Zheng Hui, Xinbo Gao, Yunchu Yang, and Xiumei Wang.
\newblock Lightweight image super-resolution with information multi-distillation network.
\newblock In {\em ACM MM}, pages 2024--2032, 2019.

\bibitem{LAPAR}
Wenbo Li, Kun Zhou, Lu~Qi, Nianjuan Jiang, Jiangbo Lu, and Jiaya Jia.
\newblock {LAPAR:} linearly-assembled pixel-adaptive regression network for single image super-resolution and beyond.
\newblock In {\em NeurIPS}, volume~33, 2020.

\bibitem{SMSR}
Longguang Wang, Xiaoyu Dong, Yingqian Wang, Xinyi Ying, Zaiping Lin, Wei An, and Yulan Guo.
\newblock Exploring sparsity in image super-resolution for efficient inference.
\newblock In {\em {CVPR}}, pages 4917--4926, 2021.

\bibitem{PAN}
Hengyuan Zhao, Xiangtao Kong, Jingwen He, Yu~Qiao, and Chao Dong.
\newblock Efficient image super-resolution using pixel attention.
\newblock In {\em {ECCVW}}, pages 56--72, 2020.

\bibitem{ECBSR}
Xindong Zhang, Hui Zeng, and Lei Zhang.
\newblock Edge-oriented convolution block for real-time super-resolution on mobile devices.
\newblock In {\em ACM MM}, pages 4034--4043, 2021.

\bibitem{FDIWN}
Guangwei Gao, Wenjie Li, Juncheng Li, Fei Wu, Huimin Lu, and Yi~Yu.
\newblock Feature distillation interaction weighting network for lightweight image super-resolution.
\newblock In {\em {AAAI}}, pages 661--669, 2022.

\bibitem{shufflemixer}
Long Sun, Jinshan Pan, and Jinhui Tang.
\newblock Shufflemixer: An efficient convnet for image super-resolution.
\newblock {\em NeurIPS}, 35:17314--17326, 2022.

\bibitem{ViT}
Alexey Dosovitskiy, Lucas Beyer, Alexander Kolesnikov, and et~al.
\newblock An image is worth 16x16 words: Transformers for image recognition at scale.
\newblock In {\em {ICLR}}, 2021.

\bibitem{SwinT}
Ze~Liu, Yutong Lin, Yue Cao, Han Hu, Yixuan Wei, Zheng Zhang, Stephen Lin, and Baining Guo.
\newblock Swin transformer: Hierarchical vision transformer using shifted windows.
\newblock In {\em {ICCV}}, 2021.

\bibitem{EDT}
Wenbo Li, Xin Lu, Shengju Qian, and Jiangbo Lu.
\newblock On efficient transformer-based image pre-training for low-level vision.

\bibitem{SwinIR}
Jingyun Liang, Jiezhang Cao, Guolei Sun, Kai Zhang, Luc~Van Gool, and Radu Timofte.
\newblock {SwinIR}: Image restoration using swin transformer.
\newblock In {\em ICCV Workshops}, pages 1833--1844, 2021.

\bibitem{ELAN}
Xindong Zhang, Hui Zeng, Shi Guo, and Lei Zhang.
\newblock Efficient long-range attention network for image super-resolution.
\newblock In {\em ECCV}, pages 649--667, 2022.

\bibitem{HAT}
Xiangyu Chen, Xintao Wang, Jiantao Zhou, Yu~Qiao, and Chao Dong.
\newblock Activating more pixels in image super-resolution transformer.
\newblock In {\em CVPR}, pages 22367--22377, 2023.

\bibitem{CAT}
Zheng Chen, Yulun Zhang, Jinjin Gu, Yongbing Zhang, Linghe Kong, and Xin Yuan.
\newblock Cross aggregation transformer for image restoration.
\newblock In {\em NeurIPS}, 2022.

\bibitem{GRL}
Yawei Li, Yuchen Fan, Xiaoyu Xiang, Denis Demandolx, Rakesh Ranjan, Radu Timofte, and Luc~Van Gool.
\newblock Efficient and explicit modelling of image hierarchies for image restoration.
\newblock In {\em CVPR}, pages 18278--18289, 2023.

\bibitem{MetaFormer}
Weihao Yu, Mi~Luo, Pan Zhou, Chenyang Si, Yichen Zhou, Xinchao Wang, Jiashi Feng, and Shuicheng Yan.
\newblock Metaformer is actually what you need for vision.
\newblock In {\em CVPR}, pages 10809--10819, 2022.

\bibitem{Conv2Former}
Qibin Hou, Cheng-Ze Lu, Ming-Ming Cheng, and Jiashi Feng.
\newblock Conv2former: A simple transformer-style convnet for visual recognition.
\newblock {\em IEEE Transactions on Pattern Analysis and Machine Intelligence}, pages 1--10, 2024.

\bibitem{VAN}
Meng{-}Hao Guo, Cheng{-}Ze Lu, Zheng{-}Ning Liu, Ming{-}Ming Cheng, and Shi{-}Min Hu.
\newblock Visual attention network.
\newblock {\em Comput. Vis. Media}, 9(4):733--752, 2023.

\bibitem{largekernel}
Xiaohan Ding, Xiangyu Zhang, Yizhuang Zhou, Jungong Han, Guiguang Ding, and Jian Sun.
\newblock Scaling up your kernels to 31x31: Revisiting large kernel design in cnns.
\newblock In {\em CVPR}, pages 11953--11965, 2022.

\bibitem{convnet}
Zhuang Liu, Hanzi Mao, Chao{-}Yuan Wu, Christoph Feichtenhofer, Trevor Darrell, and Saining Xie.
\newblock A convnet for the 2020s.
\newblock {\em {CVPR}}, pages 11966--11976, 2022.

\bibitem{Uformer}
Zhendong Wang, Xiaodong Cun, Jianmin Bao, Wengang Zhou, Jianzhuang Liu, and Houqiang Li.
\newblock Uformer: {A} general u-shaped transformer for image restoration.
\newblock In {\em CVPR}, pages 17662--17672, 2022.

\bibitem{Restormer}
Syed~Waqas Zamir, Aditya Arora, Salman Khan, Munawar Hayat, Fahad~Shahbaz Khan, and Ming{-}Hsuan Yang.
\newblock Restormer: Efficient transformer for high-resolution image restoration.
\newblock In {\em CVPR}, pages 5718--5729, 2022.

\bibitem{RepVGG}
Xiaohan Ding, Xiangyu Zhang, Ningning Ma, Jungong Han, Guiguang Ding, and Jian Sun.
\newblock Repvgg: Making vgg-style convnets great again.
\newblock In {\em CVPR}, pages 13733--13742, 2021.

\bibitem{ACNet}
Xiaohan Ding, Yuchen Guo, Guiguang Ding, and Jungong Han.
\newblock Acnet: Strengthening the kernel skeletons for powerful {CNN} via asymmetric convolution blocks.
\newblock In {\em {ICCV}}, pages 1911--1920, 2019.

\bibitem{survey}
Longlong Jing and Yingli Tian.
\newblock Self-supervised visual feature learning with deep neural networks: {A} survey.
\newblock {\em {IEEE} Trans. Pattern Anal. Mach. Intell.}, 43(11):4037--4058, 2021.

\bibitem{ACMComputingSurvey}
Saeed Anwar, Salman~H. Khan, and Nick Barnes.
\newblock A deep journey into super-resolution: {A} survey.
\newblock {\em {ACM} Comput. Surv.}, 53(3):60:1--60:34, 2021.

\bibitem{RDN}
Yulun Zhang, Yapeng Tian, Yu~Kong, Bineng Zhong, and Yun Fu.
\newblock Residual dense network for image super-resolution.
\newblock In {\em {CVPR}}, pages 2472--2481, 2018.

\bibitem{DBPN}
Muhammad Haris, Gregory Shakhnarovich, and Norimichi Ukita.
\newblock Deep back-projection networks for super-resolution.
\newblock In {\em {CVPR}}, pages 1664--1673, 2018.

\bibitem{DRRN}
Ying Tai, Jian Yang, and Xiaoming Liu.
\newblock Image super-resolution via deep recursive residual network.
\newblock In {\em {CVPR}}, pages 2790--2798, 2017.

\bibitem{channel_attention}
Jie Hu, Li~Shen, Samuel Albanie, Gang Sun, and Enhua Wu.
\newblock Squeeze-and-excitation networks.
\newblock volume~42, pages 2011--2023, 2020.

\bibitem{IDN}
Zheng Hui, Xiumei Wang, and Xinbo Gao.
\newblock Fast and accurate single image super-resolution via information distillation network.
\newblock In {\em {CVPR}}, pages 723--731, 2018.

\bibitem{Kong2022ResidualLF}
F.~Kong, Mingxi Li, Songwei Liu, Ding Liu, Jingwen He, Yang Bai, Fangmin Chen, and Lean Fu.
\newblock Residual local feature network for efficient super-resolution.
\newblock {\em CVPRW}, pages 765--775, 2022.

\bibitem{yu2023dipnet}
Lei Yu, Xinpeng Li, Youwei Li, Ting Jiang, Qi~Wu, Haoqiang Fan, and Shuaicheng Liu.
\newblock Dipnet: Efficiency distillation and iterative pruning for image super-resolution.
\newblock In {\em CVPRW}, pages 1692--1701, 2023.

\bibitem{wan2024swift}
Cheng Wan, Hongyuan Yu, Zhiqi Li, Yihang Chen, Yajun Zou, Yuqing Liu, Xuanwu Yin, and Kunlong Zuo.
\newblock Swift parameter-free attention network for efficient super-resolution.
\newblock In {\em CVPRW}, pages 6246--6256, 2024.

\bibitem{IPT}
Hanting Chen, Yunhe Wang, Tianyu Guo, Chang Xu, Yiping Deng, Zhenhua Liu, Siwei Ma, Chunjing Xu, Chao Xu, and Wen Gao.
\newblock Pre-trained image processing transformer.
\newblock In {\em {CVPR}}, pages 12299--12310, 2021.

\bibitem{HP_transSR}
Qing Cai, Yiming Qian, Jinxing Li, Jun Lyu, Yee-Hong Yang, Feng Wu, and David Zhang.
\newblock Hipa: Hierarchical patch transformer for single image super resolution.
\newblock {\em IEEE Trans. Image Process.}, 32:3226--3237, 2023.

\bibitem{LocalViT}
Yawei Li, Kai Zhang, Jiezhang Cao, Radu Timofte, and Luc~Van Gool.
\newblock Localvit: Bringing locality to vision transformers.
\newblock {\em CoRR}, abs/2104.05707, 2021.

\bibitem{TCSR}
Gang Wu, Junjun Jiang, Yuanchao Bai, and Xianming Liu.
\newblock Incorporating transformer designs into convolutions for lightweight image super-resolution.
\newblock {\em CoRR}, abs/2303.14324, 2023.

\bibitem{OMNI}
Hang Wang, Xuanhong Chen, Bingbing Ni, Yutian Liu, and Jinfan Liu.
\newblock Omni aggregation networks for lightweight image super-resolution.
\newblock {\em CoRR}, abs/2304.10244, 2023.

\bibitem{NGramSwin}
Haram Choi, Jeongmin Lee, and Jihoon Yang.
\newblock N-gram in swin transformers for efficient lightweight image super-resolution.
\newblock pages 2071--2081, 2023.

\bibitem{SLAK}
Shiwei Liu, Tianlong Chen, Xiaohan Chen, Xuxi Chen, Qiao Xiao, Boqian Wu, Mykola Pechenizkiy, Decebal~Constantin Mocanu, and Zhangyang Wang.
\newblock More convnets in the 2020s: Scaling up kernels beyond 51x51 using sparsity.
\newblock {\em CoRR}, abs/2207.03620, 2022.

\bibitem{ConvViT}
Haiping Wu, Bin Xiao, Noel Codella, Mengchen Liu, Xiyang Dai, Lu~Yuan, and Lei Zhang.
\newblock Cvt: Introducing convolutions to vision transformers.
\newblock In {\em ICCV}, pages 22--31, 2021.

\bibitem{RepSR}
Xintao Wang, Chao Dong, and Ying Shan.
\newblock Repsr: Training efficient vgg-style super-resolution networks with structural re-parameterization and batch normalization.
\newblock In {\em {ACM} Multimedia}, pages 2556--2564. {ACM}, 2022.

\bibitem{RepDySR}
Yan Wang, Tongtong Su, Yusen Li, Jiuwen Cao, Gang Wang, and Xiaoguang Liu.
\newblock Ddistill-sr: Reparameterized dynamic distillation network for lightweight image super-resolution.
\newblock {\em IEEE Trans. Multimedia}, pages 1--13, 2022.

\bibitem{dynamicConv}
Yinpeng Chen, Xiyang Dai, Mengchen Liu, Dongdong Chen, Lu~Yuan, and Zicheng Liu.
\newblock Dynamic convolution: Attention over convolution kernels.
\newblock In {\em {CVPR}}, pages 11027--11036, 2020.

\bibitem{fang2020soft}
Faming Fang, Juncheng Li, and Tieyong Zeng.
\newblock Soft-edge assisted network for single image super-resolution.
\newblock {\em IEEE Trans. Image Process.}, 29:4656--4668, 2020.

\bibitem{LAM}
Jinjin Gu and Chao Dong.
\newblock Interpreting super-resolution networks with local attribution maps.
\newblock In {\em {CVPR}}, pages 9199--9208, 2021.

\bibitem{DRSAN}
Karam Park, Jae~Woong Soh, and Nam~Ik Cho.
\newblock A dynamic residual self-attention network for lightweight single image super-resolution.
\newblock {\em IEEE Trans. Multimedia}, 25:907--918, 2023.

\bibitem{RFDN}
Jie Liu, Jie Tang, and Gangshan Wu.
\newblock Residual feature distillation network for lightweight image super-resolution.
\newblock In {\em ECCVW}, pages 41--55, 2020.

\bibitem{Ko_2022_CVPR}
Minsu Ko, Eunju Cha, Sungjoo Suh, Huijin Lee, Jae-Joon Han, Jinwoo Shin, and Bohyung Han.
\newblock Self-supervised dense consistency regularization for image-to-image translation.
\newblock In {\em CVPR}, pages 18301--18310, June 2022.

\bibitem{DLGSANet}
Xiang Li, Jinshan Pan, Jinhui Tang, and Jiangxin Dong.
\newblock Dlgsanet: Lightweight dynamic local and global self-attention networks for image super-resolution.
\newblock {\em CoRR}, abs/2301.02031, 2023.

\bibitem{omni_sr}
Hang Wang, Xuanhong Chen, Bingbing Ni, Yutian Liu, and Liu jinfan.
\newblock Omni aggregation networks for lightweight image super-resolution.
\newblock In {\em CVPR}, pages 22378--22387, 2023.

\bibitem{LatticeNet}
Xiaotong Luo, Yuan Xie, Yulun Zhang, Yanyun Qu, Cuihua Li, and Yun Fu.
\newblock Latticenet: Towards lightweight image super-resolution with lattice block.
\newblock In {\em ECCV}, pages 272--289, 2020.

\bibitem{DDistill}
Yan Wang, Tongtong Su, Yusen Li, Jiuwen Cao, Gang Wang, and Xiaoguang Liu.
\newblock Ddistill-sr: Reparameterized dynamic distillation network for lightweight image super-resolution.
\newblock {\em IEEE Trans. Multimedia}, pages 1--13, 2022.

\bibitem{DIV2K}
Eirikur Agustsson and Radu Timofte.
\newblock Ntire 2017 challenge on single image super-resolution: Dataset and study.
\newblock In {\em CVPR Workshops}, July 2017.

\bibitem{Set5}
Marco Bevilacqua, Aline Roumy, Christine Guillemot, and Marie{-}Line Alberi{-}Morel.
\newblock Low-complexity single-image super-resolution based on nonnegative neighbor embedding.
\newblock In {\em BMVC}, pages 1--10, 2012.

\bibitem{Set14}
Roman Zeyde, Michael Elad, and Matan Protter.
\newblock On single image scale-up using sparse-representations.
\newblock In {\em Curves and Surfaces}, volume 6920, pages 711--730. Springer, 2010.

\bibitem{B100}
David~R. Martin, Charless~C. Fowlkes, Doron Tal, and Jitendra Malik.
\newblock A database of human segmented natural images and its application to evaluating segmentation algorithms and measuring ecological statistics.
\newblock In {\em {ICCV}}, pages 416--425, 2001.

\bibitem{Urban100}
Jia{-}Bin Huang, Abhishek Singh, and Narendra Ahuja.
\newblock Single image super-resolution from transformed self-exemplars.
\newblock In {\em {CVPR}}, pages 5197--5206, 2015.

\bibitem{Manga109}
Yusuke Matsui, Kota Ito, Yuji Aramaki, Azuma Fujimoto, Toru Ogawa, Toshihiko Yamasaki, and Kiyoharu Aizawa.
\newblock Sketch-based manga retrieval using manga109 dataset.
\newblock {\em Multim. Tools Appl.}, 2017.

\bibitem{ADAM}
Diederik~P. Kingma and Jimmy Ba.
\newblock Adam: {A} method for stochastic optimization.
\newblock In {\em {ICLR}}, 2015.

\bibitem{Pytorch}
Adam Paszke, Sam Gross, Francisco Massa, et~al.
\newblock {PyTorch}: An imperative style, high-performance deep learning library.
\newblock In {\em NeurIPS}, pages 8024--8035, 2019.

\bibitem{RealSRdata}
Jianrui Cai, Huiyu Zeng, Hongwei Yong, Zisheng Cao, and Lei Zhang.
\newblock Toward real-world single image super-resolution: A new benchmark and a new model.
\newblock {\em ICCV}, pages 3086--3095, 2019.

\end{thebibliography}
\end{document}